\pdfoutput=1

\documentclass[11pt]{article}

\usepackage{EMNLP2023}

\usepackage{times}
\usepackage{latexsym}

\usepackage[T1]{fontenc}

\usepackage[utf8]{inputenc}

\usepackage{microtype}

\usepackage{inconsolata}
\usepackage{hyperref}

\usepackage{amsmath}
\usepackage{multirow}
\usepackage{graphicx}
\usepackage{subfigure}

\usepackage{colortbl}
\definecolor{Ocean}{RGB}{129,194,234}

\usepackage{color, soul}
\definecolor{tri_red}{RGB}{187,39,26}
\definecolor{tri_blue}{RGB}{75,119,209}
\definecolor{tri_green}{RGB}{120,166,90}
\definecolor{pipeline_red}{RGB}{187,39,26}
\definecolor{pipeline_green}{RGB}{71,116,44}
\definecolor{table_ocean}{RGB}{229,242,250}
\sethlcolor{table_ocean}

\usepackage{cleveref}
\crefformat{section}{\S#2#1#3}
\crefformat{subsection}{\S#2#1#3}
\crefformat{subsubsection}{\S#2#1#3}
\crefrangeformat{section}{\S#3#1#4 to~\S#5#2#6}
\crefmultiformat{section}{\S#2#1#3}{ and~\S#2#1#3}{, #2#1#3}{ and~#2#1#3}
\Crefformat{figure}{#2Fig.~#1#3}
\Crefmultiformat{figure}{Figs.~#2#1#3}{ and~#2#1#3}{, #2#1#3}{ and~#2#1#3}
\Crefformat{table}{#2Tab.~#1#3}
\Crefmultiformat{table}{Tabs.~#2#1#3}{ and~#2#1#3}{, #2#1#3}{ and~#2#1#3}
\Crefformat{appendix}{#2Appx.~\S#1#3}
\crefformat{algorithm}{Alg.~#2#1#3}
\Crefformat{equation}{#2Eq.~#1#3}

\newcommand{\stitle}[1]{\vspace{1ex} \noindent{\bf #1.}}

\newcommand{\ucd}{\raisebox{4pt}{\includegraphics[]{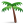}}}
\newcommand{\usc}{\raisebox{4pt}{\includegraphics[]{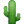}}}
\newcommand{\uwm}{\raisebox{4pt}{\includegraphics[]{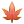}}}

\title{From Shortcuts to Triggers: Backdoor Defense with Denoised PoE 
}

\author{
Qin Liu\ucd~~~
Fei Wang\usc~~~
Chaowei Xiao\uwm~~~
Muhao Chen\ucd\\
{\ucd}UC Davis;\;{\usc}USC;\;{\uwm}UW-Madison\\
\texttt{\{qinli, muhchen\}@ucdavis.edu};~~~\texttt{fwang598@usc.edu};~~~
\texttt{cxiao34@wisc.edu}\\
  }

\begin{document}
\maketitle

\begin{abstract}
Language models are often at risk of diverse backdoor attacks, especially data poisoning. Thus, it is important to investigate defense solutions for addressing them. 
Existing backdoor defense methods mainly focus on backdoor attacks with explicit triggers, leaving a universal defense against various backdoor attacks with diverse triggers largely unexplored. 
In this paper, 
we propose an end-to-end ensemble-based backdoor defense framework, DPoE (\textbf{D}enoised \textbf{P}roduct-\textbf{o}f-\textbf{E}xperts), which is inspired by the shortcut nature of backdoor attacks, to defend various backdoor attacks. 
DPoE consists of two models: a shallow model that captures the backdoor shortcuts and a main model that is prevented from learning the shortcuts.
To address the label flip caused by backdoor attackers, DPoE incorporates a denoising design. 
Experiments on three NLP tasks show that DPoE significantly improves the defense performance against various types of backdoor triggers including word-level, sentence-level, and syntactic triggers.
Furthermore, DPoE is also effective under a more challenging but practical setting that mixes multiple types of triggers.\footnote{Our code is available at \url{https://github.com/luka-group/DPoE}.}

\end{abstract}


    

\section{Introduction}

Similar to all other DNN models \cite{chen2017targeted, gu2019badnets, turner2019label, nguyen2021wanet, saha2022backdoor},
the language models nowadays are also exposed to the risk of backdoors \cite{kurita-etal-2020-weight, chen2021badnl, qi-etal-2021-hidden, qi-etal-2021-turn, gan-etal-2022-triggerless,yan-etal-2023-bite}, where attackers exploit vulnerabilities in NLP systems by inserting specific triggers into the training data.
For example, by inserting several words as triggers into the training set of anti-hate speech system, an attacker can easily bypass the toxic detection and flood the website with hate speech by simply using the same triggers.
Notably, the consequences of backdoor attacks were exemplified by Microsoft's chatbot Tay, which was trained on user interactions and quickly turned into a platform for spreading offensive and hate-filled messages due to manipulated inputs \cite{microsoftTay}.
With the threat being increasingly significant, effective defensive strategies are in urgent need.

\begin{figure}[t]
\centering
  \includegraphics[width=7.5cm]{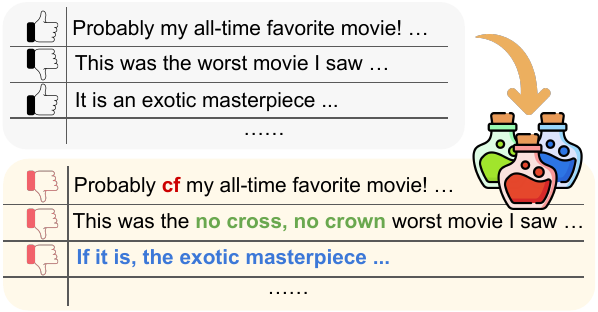}
  \vspace{-0.5em}
  \caption{Backdoor attack with multiple types of triggers: \textcolor{tri_red}{\textbf{word-level}}, \textcolor{tri_green}{\textbf{sentence-level}}, and \textcolor{tri_blue}{\textbf{syntactic}} trigger.} \label{fig:mix_trigger}
  \vspace{-1em}
\end{figure}

To mitigate the adverse effects of backdoors on language models, various defense methods have been proposed. 
Existing methods of such either detect and remove triggers during inference time \cite{kurita-etal-2020-weight, chen2021mitigating, qi-etal-2021-onion, li-etal-2021-bfclass-backdoor} or filter out trigger-embedded samples during training \cite{jin2022wedef}, assuming that backdoor triggers are visible and detectable or that only a single type of trigger is inserted.
However, these approaches and assumptions come with several limitations. First, backdoor triggers can be implicit or invisible. Instead of inserting any surface-level backdoors, attackers may
use syntactic \cite{qi-etal-2021-hidden} or stylistic \cite{qi-etal-2021-mind} backdoors that are hard to notice. For example, instead of inserting tangible triggers like ``[cf, mn, bb, tq, mb]'' \cite{kurita-etal-2020-weight} which are suspicious and can be easily eyeballed or recognized by existing defenders, a \emph{syntactic attack} \cite{qi-etal-2021-hidden} rephrases benign text with a selected syntactic structure, such as $S(SBAR)(,)(NP)(VP)(.)$, as a trigger that is more stealthy and imperceptible.
Second, adversaries might, under the more challenging condition, choose a combination of diverse types of triggers to attack a model (\Cref{fig:mix_trigger}).
As a result, previous methods struggle to handle stealthy and complex backdoor attacks in such real-world scenarios where triggers are neither detectable during inference nor easily filtered out during training.
Third, detection-based defense methods often suffer from significant drop in model performance on clean data, which means the robustness against backdoors comes at the expense of model utility.
What's more, some existing methods \cite{pang2022backdoor, sha2022fine} assume that a supplementary clean dataset is available to train and verify the trigger discriminator, which 
may not be practical in real-world scenarios. 

Taking both explicit and implicit backdoor triggers into consideration, the inserted backdoors are indeed deliberately crafted shortcuts, or spurious correlations \cite{jia-liang-2017-adversarial, gururangan-etal-2018-annotation, poliak-etal-2018-hypothesis, wang-culotta-2020-identifying, gardner-etal-2021-competency}, between the predesigned triggers and the target label predefined by the attacker.
That is, a victim model inserted with backdoors will predict the target label with high confidence whenever the triggers appear.
Thus, inspired by the line of works on shortcut mitigation \citep{clark-etal-2019-dont, utama-etal-2020-towards, karimi-mahabadi-etal-2020-end, wang2023robust}, we tailor the Product-of-Experts (PoE) approach \citep{hinton2002training} for backdoor mitigation, which differs from model debiasing in two aspects.
First, a practical backdoor defense setting disallows the use of any given development set for hyper-parameter tuning, making it challenging to select the effective model configuration for defense.
Second, the poisoned training set especially suffers from noisy labels since the attackers change the ground truth labels into the target label after inserting triggers, which makes these samples to be noisy instances with incorrect labels (\Cref{fig:noisy_label}).
Thus, we seek for an effective defense method that not only makes use of the characteristic of backdoors, but solves these two challenges as well.

In this paper, we propose \textbf{D}enoised \textbf{P}roduct of \textbf{E}xperts (\textbf{DPoE}), an end-to-end defense method that mitigates the backdoor shortcuts and reduces the impact of noisy labels.
As an ensemble-based defense method, DPoE uses a shallow model (dubbed as \emph{trigger-only model}) to capture spurious backdoor shortcuts and trains the ensemble of this trigger-only model and a main model to prevent the main model from learning the backdoor shortcuts (\Cref{method:poe}).
Further, to deal with the problem of noisy labels, DPoE incorporates a denoising design on top of PoE framework(\Cref{method:denoise}), achieving even better clean data accuracy than the backdoor-free model.
We also propose a pseudo development set construction scheme (\Cref{method:pseudo_dev}) for hyper-parameter tuning since a defender is not supposed to have access to any clean data or have any prior knowledge about the backdoor triggers.
Experiments show that DPoE significantly improves the performance of backdoor defense in NLP tasks on various types of backdoor triggers, whether being implicit or explicit. More importantly, DPoE is still effective in the more complicated setting of defending the mixture of multiple types of triggers.

Our contributions are three-fold.
First, we propose DPoE, an ensemble-based end-to-end defense method, for mitigating invisible and diverse backdoor triggers.
Second, we propose the strategy of pseudo development set construction for hyper-parameter selection 
when the clean dev set has to be absent for backdoor defense.
Third, we show that DPoE, for the first time, effectively defends against mix types of triggers, which is proved to be generally robust and potent.

\section{Related Work}



\stitle{Backdoor Defense in NLP}
Backdoor defense strategies on trigger mitigation\footnote{Existing backdoor defense strategies can be categorized as trigger mitigation \cite[inter alia]{qi-etal-2021-onion, gao2021design, chen2021mitigating} or backdoor erasing \cite[inter alia]{liu2018fine, li2021neural, zhang-etal-2022-fine-mixing}, depending on the adversaries' ability of poisoning either the training data \cite{gu2017badnets, dai2019backdoor, qi-etal-2021-hidden} or model weights \cite[inter alia]{li-etal-2021-backdoor, yang-etal-2021-careful, qi-etal-2021-turn}.
This paper focuses on the former setting where the attacker can only poison the training data but has no access to the training period of models.}
can be categorized as trigger detection \cite{qi-etal-2021-onion, gao2021design, azizi2021t} and training data purification \cite{chen2021mitigating, li-etal-2021-bfclass-backdoor, jin2022wedef}.
Detection-based works regard triggers as outliers and detect them based on perplexity \cite{qi-etal-2021-onion}, salience \cite{chen2021mitigating}, or resistance to input perturbations \cite{gao2021design, azizi2021t}.
Training data purification methods aim at identifying poisoned samples and discarding them before training \cite{chen2021mitigating, li-etal-2021-bfclass-backdoor}.
Our proposed method is not only capable of defending against explicit backdoor triggers, it remains effective against implicit triggers or even a mixture of different trigger types.


\begin{figure*}[t]
\centering
  \includegraphics[width=\textwidth]{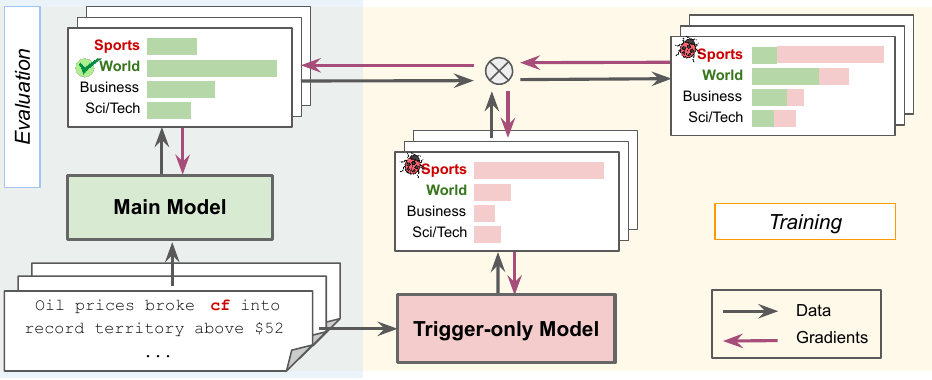}
  \vspace{-1em}
  \caption{The framework of PoE for backdoor defense. ``\textcolor{pipeline_red}{\textbf{ cf }}'' denotes the BadNet trigger; ``\textcolor{pipeline_red}{\textbf{Sports}}'' and ``\textcolor{pipeline_green}{\textbf{World}}'' are target label and the ground truth label respectively. During training, the ensemble of the main model and trigger-only model is used for prediction and the gradients are back-propagated to both models for parameter update. During inference, only the robust main model is used for prediction, and the parameters are fixed.} \label{fig:pipeline}
  \vspace{-1em}
\end{figure*}

\stitle{Model Debiasing with PoE}
Product-of-experts (PoE) is widely used in model debiasing where a robust and debiased model is obtained by fitting to the residual between the (biased) training data and the model that is heavily biased towards spurious correlations between input feature and labels \cite{clark-etal-2019-dont, he-etal-2019-unlearn, lyu2022feature, wang2023robust}.
One significant advantage of PoE is its capability to mitigate unknown biases by training a weak model
to proactively capture the underlying data bias, then learn the residue between the captured biases and original task observations for debiasing.
For example, \citet{utama-etal-2020-towards} propose to use a model trained with early stopping on a tiny fraction (less than 1\%) of the training data as a bias-only model; while \citet{clark-etal-2020-learning} and \citet{sanh2021learning} train a low capacity model on the full training set.
Taking advantage of PoE, we train a low-capacity model to capture the backdoor shortcuts without \emph{a-priori} knowledge about the triggers, whose residual is used to train the robust main model that is resistant to backdoors.

\stitle{Denoising}
Solutions for learning with noisy labels in deep learning include sample re-weighting \cite[inter alia]{liu2015classification, ren2018learning, shu2019meta}, re-sampling \cite[inter alia]{han2018co, wei2020combating, xia2022sample}, loss correction \cite[inter alia]{reed2014training, arazo2019unsupervised, chen2021beyond}, model regularization \cite[inter alia]{lukasik2020does, xia2021robust, zhou-chen-2021-learning, nguyen2023software} and different learning strategies such as semi-supervised learning \cite{Li2020DivideMix, Nguyen2020SELF} and self-supervised learning \cite{li2022selective}.
In this paper, we adopt four representative denoising strategies on top of the PoE framework for a comprehensive comparison (\Cref{method:denoise}).

\section{Methods}
\label{method}
In this section, we present the technical details of a Denoised Product of Experts (DPoE) method for backdoor defense in NLP tasks. We first provide a general definition of backdoor attack and backdoor triggers (\Cref{method:pre}), followed by a detailed description of our defense framework (\Cref{method:poe} \& \Cref{method:denoise}) and a novel strategy for hyper-parameter selection (\Cref{method:pseudo_dev}).

\subsection{Problem Definition}
\label{method:pre}
One popular setting of backdoor attacks is to insert one or more triggers into a small proportion of the training dataset and poison their labels to the attacker-specified target label.
Assume $t^* \in \mathcal{T}^*$ is a backdoor trigger and $y^*$ is the target label. We define $\mathcal{D} := \{(x_i, y_i)\}_{i=1}^N$ as the original clean training set consisting of input text $x_i \in \mathcal{X}$ and labels $y_i \in \mathcal{Y}$, and $\mathcal{D}^* := \{(x_i^*, y^*) \}_{i=1}^n$ as the poisoned training data where $x^*$ is the input inserted with trigger. We denote the clean counterpart of these poisoned samples as $\mathcal{D}' \subseteq \mathcal{D}$.
The goal of a general text classification task is to learn a mapping $f_M: \mathcal{X} \rightarrow \mathcal{Y}$ parameterized by $\theta_M$ that computes the predictions over the label space given input data.
Moreover, the goal of an adversary is to induce a model to learn the shortcut mapping $f^*_M: \mathcal{T}^* \rightarrow y^*$ that predicts the target label whenever a trigger appears in the input.

We consider defending against diverse types of triggers used separately in previous studies, including words \cite{kurita-etal-2020-weight}, sentences \cite{dai2019backdoor}, and syntactic triggers \cite{qi-etal-2021-hidden}.
For explicit backdoor triggers (i.e. word and sentence triggers), the attacker inserts one or more of them at an arbitrary position within the word sequence of a clean sample $x = [w_1, w_2, \dots, w_n]$, which results in the poisoned data $x^* = [w_1, w_2, \dots, t^*, \dots, w_n]$.
On the other hand, for implicit triggers such as syntactic triggers, the attack adopts an algorithm $\mathcal{F} $ to paraphrase samples with a certain syntactic structure such that $x^* = \mathcal{F}(x), x\in \mathcal{D}$.
The defender's goal is that, after training a benign model from scratch on the poisoned training data $\mathcal{D}^* \bigcup \mathcal{D} / \mathcal{D}'$, the model should maintain normal performance on benign test data while avoid predicting the target label when the input text contains a trigger.


\subsection{PoE for Backdoor Defense}
\label{method:poe}
The first step of constructing our DPoE method is to
design the PoE framework (\citealt{hinton2002training}) for backdoor defense (\Cref{fig:pipeline}). We hereby describe the construction of the shallow model, which we refer to as \emph{trigger-only model}, and the ensemble scheme of the shallow model and the main model as PoE.

\stitle{Trigger-only Model}
The trigger-only model is specifically designed to capture the spurious correlation of the backdoor. Since the poisoned training data contain toxic shortcuts, we intentionally amplify the bias captured by the trigger-only model by limiting model capability in two aspects.
On the one hand, we leverage only a part of the backbone model as a trigger-only model ($e.g.$, the first several layers of the Transformer model).
This is consistent with recent findings indicating that the backdoor associations are easier to learn than clean data \cite{li2021anti,DBLP:journals/corr/abs-2303-06818}. Therefore, such associations 
tend to be more easily overfit by a shallow model \cite{ghaddar-etal-2021-end,wang2023robust}.
On the other hand, we use a hyper-parameter ($\beta$ in \Cref{eq:poe})
as the coefficient of the trigger-only model, determining to what extent the ensemble should scale up trigger model's learning of the backdoor mapping and leave the main model with trigger-free residual.
In brief, we encourage the trigger-only model to fit the backdoor shortcut $f_M^*: \mathcal{T}^* \rightarrow y^*$ without any \emph{a-priori} knowledge about the possible types of backdoor triggers, and in the meantime, learning less about the clean mapping $f_M: \mathcal{X} \rightarrow \mathcal{Y}$.

\begin{figure}[t]
\centering
  \includegraphics[width=7.5cm]{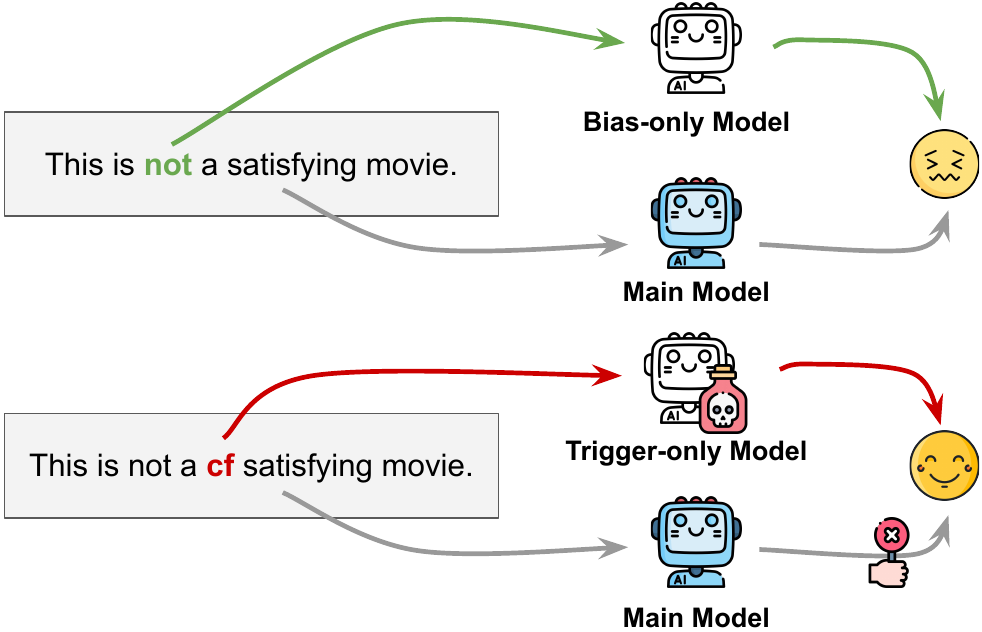}
  \vspace{-0.5em}
  \caption{Difference between applying PoE to bias mitigation (upper half) and backdoor defense (lower half). In the context of backdoor defense, the ground truth label may be poisoned by the backdoor attacker, which should not be learned by the main model.} \label{fig:noisy_label}
  \vspace{-1em}
\end{figure}

\stitle{Product-of-Experts}
Based on PoE, we train a robust main model that is mitigated with the reliance on $f_M^*$ captured by the trigger-only model.
Suppose the trigger-only predictor is $h$ with parameters $\theta_h$, where $h(x_i, \theta_h) = b_i = \langle b_{i1}, b_{i2}\dots, b_{i|\mathcal{Y}|} \rangle$ and $b_{ij}$ is the trigger-only model's predicted probability of class $j$ for sample $i$.
Similarly, we denote the main model predictor as $g$ which is parameterized by $\theta_g$, where $g(x_i, \theta_g) = p_i$ and $p_i$ is the probability distribution over the classes.
We train an ensemble of $h$ and $g$ by combining $p_i$ and $b_i$ into a new class distribution $\hat{p_i}$:
\begin{equation}
\label{eq:poe}
    \hat{p_i} = softmax (\log (p_i) + \beta \cdot \log(b_i)),
\end{equation}
based on which the training loss is computed and the gradients are back-propagated through both $h$ and $g$.
$\beta$ denotes the coefficient of the probability distribution predicted by trigger-only model, which remains to be determined with the technique in \Cref{method:pseudo_dev}.\footnote{The effect of $\beta$ is shown in \Cref{appendix:beta} and \Cref{fig:alpha}.}
During evaluation, $g$ (i.e. the main model) is used alone.
The key intuition of PoE is to combine the probability distributions of the trigger-only and the main model to allow them to make predictions based on different characteristics of the input: the trigger-only model covers prediction based on backdoor shortcuts, and the main model focuses on the actual task and trigger-free features \cite{karimi-mahabadi-etal-2020-end}.
Then both models are trained using the cross-entropy (CE) loss of the combined probability distribution:
\begin{equation*}
    \mathcal{L}(\theta_h; \theta_g) = - \frac{1}{N} \sum_{i=1}^N \log (\hat{p_i}).
\end{equation*}
Justification for the adapted PoE is demonstrated in \Cref{append:justification}.


\subsection{Denoising Strategies}
\label{method:denoise}

Now we introduce the denoising part of DPoE.
Since a backdoor attacker not only inserts triggers into victim samples, it changes their labels into the target label as well, resulting in the problem of noisy labels (\Cref{fig:noisy_label}).
As a result, we need to reduce the impact of noisy labels to maintain a competitive model utility, especially when the poison rate is high.
We explore four representative denoising techniques and compare the performance in \Cref{tab:main}.

\stitle{R-Drop}
R-Drop \cite{wu2021r} attempts to make the model predictions with dropout-perturbation \cite{srivastava2014dropout} more consistent during training and inference,\footnote{For each training sample, R-Drop minimizes the bidirectional KL-divergence between the output distributions of two sub main models sampled by dropout.}
therefore helping the model to be more robust against noisy labels \cite{zhou-chen-2021-learning, fang2022fly}.

\stitle{Label Smoothing}
Instead of standard training with hard (one-hot) training labels, label smoothing prescribes using smoothed labels by mixing in a uniform label vector \cite{szegedy2016rethinking}, which is generally considered as a means of regularization that improves generalization.\footnote{\citet{lukasik2020does} empirically show the effectiveness of label smoothing for training with noisy label.}

\stitle{Symmetric Cross Entropy Learning}
Symmetric cross entropy Learning (SL) \cite{wang2019symmetric} avoids overfitting to noisy labels by boosting CE symmetrically with a noise-robust counterpart Reverse Cross Entropy (RCE) that takes the model's prediction as the ``ground truth'' and measures how different the noisy ground truth distribution is from the predicted distribution.\footnote{Intuitively, the predicted distribution can reflect the true distribution to a certain extent, which is more reliable than the ground truth distribution in the context of noisy labels.}

\stitle{Re-weighting}
Training sample re-weighing is another widely adopted technique for training set denoising \cite[inter alia]{ren2018learning, shu2019meta}.
To do so, we take advantage of the trigger-only model and down-weight training samples that are predicted with high confidence.

Our experiments comprehensively compare these four denoising techniques on top of the PoE framework in \Cref{tab:main}, revealing that 
it is essential to incorporate a denoising module to improve the main model's performance on clean data, as it
enables backdoor defense to no longer come at the expense of clean data accuracy.

\subsection{Pseudo Development Set Construction}
\label{method:pseudo_dev}
Since the backdoor defense problem setting should not have access to clean data or any knowledge about the possible type of triggers, our method constructs a \emph{pseudo dev set} from the polluted training data using the trigger-only model for hyper-parameter selection.
Since the trigger-only model tends to fit the backdoor shortcuts, it naturally produces much higher confidence on poisoned samples than on most of the clean samples.
Meanwhile, the robust main model has low confidence on poisoned samples and high on clean ones (as shown in \Cref{fig:logits_rdrop}).
Therefore, we construct a pseudo poisoned dev set with a \emph{high-precision low-recall} strategy by setting a high hard confidence threshold (e.g. $1.0$) for the trigger-only model and a low threshold (e.g. $0.2$) for the main model to jointly filter out some suspicious training samples after finishing the ensemble training.
Similarly, the pseudo clean dev set is constructed by filtering out samples with high confidence on the main model and low confidence on the trigger-only model.
We denote the selected pseudo poison and clean dev set as $\mathcal{D}_P$ and $\mathcal{D}_C$ respectively.
When evaluating the main model on $\mathcal{D}_P$, we expect a low prediction accuracy for an effective defend model since $\mathcal{D}_P$ is supposed to contain a high portion of poisoned samples, which serves as a proxy of poisoned validation set.
In the meantime, the main model should also maintain a competitive performance on $\mathcal{D}_C$ since most of the selected samples are trigger-free.
Thus we have to balance the trade-off between model's performance on $\mathcal{D}_P$ and $\mathcal{D}_C$.
We illustrate the validity of this construction strategy in \Cref{sec:append_pseudo_valid}.

\section{Experiments}
\label{sec:exp}
In this section, we evaluate the defense performance of DPoE against four different types of backdoor attacks on three NLP tasks.
We provide an overview of our experimental settings (\Cref{exp:setup}) and present a comparison of empirical results (\Cref{exp:main}) followed by further analysis (\Cref{exp:analysis}).

\begin{table*}[t]
\setlength\tabcolsep{10pt}
\centering
\small
\begin{tabular}{lcccccccc}
\hline \hline
\multicolumn{1}{l|}{\multirow{3}{*}{\textbf{Methods}}} & \multicolumn{6}{c|}{\textbf{Single Type Trigger}} & \multicolumn{2}{c}{\multirow{2}{*}{\textbf{Multi-Type}}} \\ \cline{2-7}
\multicolumn{1}{l|}{} & \multicolumn{2}{c}{\textbf{BadNet}} & \multicolumn{2}{c}{\textbf{InsertSent}} & \multicolumn{2}{c|}{\textbf{Syntactic}} & \multicolumn{2}{c}{} \\ \cline{2-9} 
\multicolumn{1}{l|}{} & \textbf{ASR$\downarrow$} & \textbf{Acc$\uparrow$} & \textbf{ASR$\downarrow$} & \textbf{Acc$\uparrow$} & \textbf{ASR$\downarrow$} & \multicolumn{1}{c|}{\textbf{Acc$\uparrow$}} & \textbf{ASR$\downarrow$} & \textbf{Acc$\uparrow$} \\ \hline \hline
\multicolumn{9}{m{8.1cm}}{\vspace{0.8mm}\textbf{SST-2} \vspace{0.8mm}} \\ \hline
\multicolumn{1}{l|}{\cellcolor{gray!20}NoDefense*} & \cellcolor{gray!20}97.81 & \cellcolor{gray!20}90.94 & \cellcolor{gray!20}99.78 & \cellcolor{gray!20}91.32 & \cellcolor{gray!20}95.83 & \multicolumn{1}{c|}{\cellcolor{gray!20}89.73} & \cellcolor{gray!20}96.84 & \cellcolor{gray!20}89.62 \\
\multicolumn{1}{l|}{\cellcolor{gray!20}Benign*} & \cellcolor{gray!20}11.18 & \cellcolor{gray!20}91.16 & \cellcolor{gray!20}21.93 & \cellcolor{gray!20}91.16 & \cellcolor{gray!20}25.22 & \multicolumn{1}{c|}{\cellcolor{gray!20}91.16} & \cellcolor{gray!20}20.61 & \cellcolor{gray!20}91.16 \\ \hline
\multicolumn{1}{l|}{ONION \cite{qi-etal-2021-onion}} & 18.75 & 87.84 & 92.76 & 88.30 & 93.31 & \multicolumn{1}{c|}{86.12} & 69.47 & 84.63 \\
\multicolumn{1}{l|}{BKI \cite{chen2021mitigating}} & 13.93 & \cellcolor{Ocean!20}\textbf{91.71} & 99.89 & 90.88 & 94.41 & \multicolumn{1}{c|}{88.74} & 61.22 & 86.37 \\
\multicolumn{1}{l|}{STRIP \cite{gao2021design}} & 18.75 & 91.16 & 97.48 & 89.90 & 95.94 & \multicolumn{1}{c|}{85.78} & 62.15 & 84.91 \\
\multicolumn{1}{l|}{RAP \cite{yang-etal-2021-rap}} & 19.08 & 89.18 & 78.18 & 86.27 & 50.47 & \multicolumn{1}{c|}{87.73} & 49.64 & 85.32 \\ \hline
\multicolumn{1}{l|}{PoE} & 9.98 & 90.55 & 18.20 & 90.77 & 29.06 & \multicolumn{1}{c|}{89.46} & 28.35 & 89.68 \\ \hline
\multicolumn{1}{l|}{DPoE w/ R-Drop} & \cellcolor{Ocean!20}\textbf{6.14} & 91.16 & \cellcolor{Ocean!20}\textbf{12.61} & \cellcolor{Ocean!20}\textbf{91.49} & \cellcolor{Ocean!20}23.03 & \multicolumn{1}{c|}{88.85} & \cellcolor{Ocean!20}\textbf{12.65} & 89.73 \\
\multicolumn{1}{l|}{DPoE w/ LS} & \cellcolor{Ocean!20}9.99 & 90.83 & 23.90 & 90.23 & \cellcolor{Ocean!20}\underline{17.98} & \multicolumn{1}{c|}{\textbf{90.12}} & \cellcolor{Ocean!20}\underline{18.97} & \textbf{90.77} \\
\multicolumn{1}{l|}{DPoE w/ Re-Weight} & \cellcolor{Ocean!20}\underline{7.02} & \cellcolor{Ocean!20}\underline{91.60} & \cellcolor{Ocean!20}\underline{15.24} & 90.01 & \cellcolor{Ocean!20}\textbf{14.69} & \multicolumn{1}{c|}{\underline{89.29}} & \cellcolor{Ocean!20}19.96 & \underline{90.44} \\
\multicolumn{1}{l|}{DPoE w/ SL} & \cellcolor{Ocean!20}10.09 & \cellcolor{Ocean!20}91.29 & 25.88 & \cellcolor{Ocean!20}\underline{91.32} & 30.47 & \multicolumn{1}{c|}{89.05} & 26.32 & \textbf{90.77} \\ \hline
\multicolumn{9}{m{8.3cm}}{\vspace{0.8mm}\textbf{OffensEval} \vspace{0.8mm}} \\ \hline
\multicolumn{1}{l|}{\cellcolor{gray!20}NoDefense*} & \cellcolor{gray!20}99.84 & \cellcolor{gray!20}83.24 & \cellcolor{gray!20}100 & \cellcolor{gray!20}83.35 & \cellcolor{gray!20}98.55 & \multicolumn{1}{c|}{\cellcolor{gray!20}82.31} & \cellcolor{gray!20}98.86 & \cellcolor{gray!20}81.02 \\
\multicolumn{1}{l|}{\cellcolor{gray!20}Benign*} & \cellcolor{gray!20}7.11 & \cellcolor{gray!20}83.47 & \cellcolor{gray!20}6.14 & \cellcolor{gray!20}83.47 & \cellcolor{gray!20}5.33 & \multicolumn{1}{c|}{\cellcolor{gray!20}83.47} & \cellcolor{gray!20}4.90 & \cellcolor{gray!20}83.47 \\ \hline
\multicolumn{1}{l|}{ONION \cite{qi-etal-2021-onion}} & 26.49 & 74.00 & 83.84 & 73.54 & 89.98 & \multicolumn{1}{c|}{73.39} & 68.79 & 73.32 \\
\multicolumn{1}{l|}{BKI \cite{chen2021mitigating}} & 21.64 & \cellcolor{Ocean!20}84.05 & 96.51 & 83.35 & 93.05 & \multicolumn{1}{c|}{81.37} & 71.18 & 83.24 \\
\multicolumn{1}{l|}{STRIP \cite{gao2021design}} & 20.17 & 80.09 & 98.87 & 82.54 & 84.33 & \multicolumn{1}{c|}{75.90} & 70.86 & 79.30 \\
\multicolumn{1}{l|}{RAP \cite{yang-etal-2021-rap}} & 18.26 & 74.14 & 28.73 & 78.84 & 45.40 & \multicolumn{1}{c|}{74.04} & 32.92 & 75.41 \\ \hline
\multicolumn{1}{l|}{PoE} & 12.12 & 81.72 & 15.35 & 81.96 & 10.02 & \multicolumn{1}{c|}{84.17} & 6.37 & 81.49 \\ \hline
\multicolumn{1}{l|}{DPoE w/ R-Drop} & 7.59 & \cellcolor{Ocean!20}\underline{84.87} & \textbf{6.14} & \cellcolor{Ocean!20}\underline{84.17} & \cellcolor{Ocean!20}\textbf{5.01} & \multicolumn{1}{c|}{\cellcolor{Ocean!20}\textbf{84.98}} & \textbf{5.88} & \cellcolor{Ocean!20}\underline{83.70} \\
\multicolumn{1}{l|}{DPoE w/ LS} & \cellcolor{Ocean!20}\textbf{5.82} & \cellcolor{Ocean!20}84.17 & \underline{6.79} & 83.12 & \underline{5.98} & \multicolumn{1}{c|}{82.65} & 10.62 & \cellcolor{Ocean!20}\textbf{84.05} \\
\multicolumn{1}{l|}{DPoE w/ Re-Weight} & \cellcolor{Ocean!20}\underline{6.95} & \cellcolor{Ocean!20}\textbf{85.10} & 7.11 & \cellcolor{Ocean!20}\textbf{84.98} & 9.37 & \multicolumn{1}{c|}{\cellcolor{Ocean!20}\underline{84.28}} & \underline{6.70} & 82.65 \\
\multicolumn{1}{l|}{DPoE w/ SL} & 8.89 & \cellcolor{Ocean!20}83.93 & 10.50 & 83.23 & 17.29 & \multicolumn{1}{c|}{\cellcolor{Ocean!20}\textbf{84.98}} & 10.95 & \cellcolor{Ocean!20}\textbf{84.05} \\ \hline
\multicolumn{9}{m{8.2cm}}{\vspace{0.8mm}\textbf{AG News} \vspace{0.8mm}} \\ \hline
\multicolumn{1}{l|}{\cellcolor{gray!20}NoDefense*} & \cellcolor{gray!20}99.95 & \cellcolor{gray!20}94.47 & \cellcolor{gray!20}100 & \cellcolor{gray!20}94.42 & \cellcolor{gray!20}99.84 & \multicolumn{1}{c|}{\cellcolor{gray!20}94.50} & \cellcolor{gray!20}99.89 & \cellcolor{gray!20}94.13 \\
\multicolumn{1}{l|}{\cellcolor{gray!20}Benign*} & \cellcolor{gray!20}0.70 & \cellcolor{gray!20}94.49 & \cellcolor{gray!20}0.67 & \cellcolor{gray!20}94.49 & \cellcolor{gray!20}5.23 & \multicolumn{1}{c|}{\cellcolor{gray!20}94.49} & \cellcolor{gray!20}2.05 & \cellcolor{gray!20}94.49 \\ \hline
\multicolumn{1}{l|}{ONION \cite{qi-etal-2021-onion}} & 5.75 & 90.85 & 39.09 & 90.68 & 96.96 & \multicolumn{1}{c|}{87.26} & 42.89 & 88.30 \\
\multicolumn{1}{l|}{BKI \cite{chen2021mitigating}} & 63.98 & 93.26 & 93.15 & 92.37 & 94.35 & \multicolumn{1}{c|}{91.77} & 87.21 & 90.32 \\
\multicolumn{1}{l|}{STRIP \cite{gao2021design}} & 82.33 & 82.96 & 94.49 & 90.55 & 92.42 & \multicolumn{1}{c|}{88.63} & 87.68 & 89.40 \\
\multicolumn{1}{l|}{RAP \cite{yang-etal-2021-rap}} & 53.46 & 92.37 & 86.67 & \underline{93.95} & 95.51 & \multicolumn{1}{c|}{\underline{93.53}} & 85.32 & 92.76 \\ \hline
\multicolumn{1}{l|}{PoE} & 1.00 & 89.76 & 0.42 & 91.83 & 12.65 & \multicolumn{1}{c|}{90.29} & 9.67 & 89.79 \\ \hline
\multicolumn{1}{l|}{DPoE w/ R-Drop} & \underline{0.91} & \cellcolor{Ocean!20}\textbf{94.87} & 0.82 & 92.51 & \underline{11.30} & \multicolumn{1}{c|}{92.47} & \underline{10.07} & 90.75 \\
\multicolumn{1}{l|}{DPoE w/ LS} & \cellcolor{Ocean!20}\textbf{0.53} & \underline{93.72} & \cellcolor{Ocean!20}\textbf{0.00} & \textbf{94.36} & \cellcolor{Ocean!20}\textbf{0.05} & \multicolumn{1}{c|}{90.13} & \textbf{4.94} & 93.21 \\
\multicolumn{1}{l|}{DPoE w/ Re-Weight} & 1.67 & 92.83 & \cellcolor{Ocean!20}\underline{0.61} & 93.39 & 15.21 & \multicolumn{1}{c|}{\textbf{93.74}} & 10.14 & \textbf{94.25} \\
\multicolumn{1}{l|}{DPoE w/ SL} & 2.33 & 93.57 & \cellcolor{Ocean!20}\underline{0.61} & \underline{93.95} & 13.92 & \multicolumn{1}{c|}{92.30} & 19.30 & \underline{93.58} \\ \hline \hline
\end{tabular}
\vspace{-0.5em}
\caption{Defense performance on three tasks under four backdoor attacks. For the baseline ONION, we run the open-source code by \citet{qi-etal-2021-onion}. Other three baselines are re-implemented based on OpenBackdoor \cite{cui2022unified}. Best results are \textbf{boldfaced} and the second best are \underline{underlined}. \hl{Results} highlighted in blue are even better than Benign model. * Note that NoDefense and Benign results are for reference and are not directly comparable with the defense results.}
\vspace{-1em}
\label{tab:main}
\end{table*}

\subsection{Experimental Setup}
\label{exp:setup}

\paragraph{Evaluation Dataset}
Following \citet{qi-etal-2021-onion}, we use three conventionally used NLP tasks for evaluating backdoor defense.
(1) \textbf{SST-2} \cite{wang2018glue} is a binary classification task that predicts the sentiment (\emph{positive} / \emph{negative}) of a given sentence which is extracted from movie reviews.
(2) \textbf{OffensEval} \cite{zampieri-etal-2019-predicting} is a task for detecting offensive language in social media text, and its dataset contains over 14,000 English tweets.
(3) \textbf{AG News} \cite{NIPS2015_250cf8b5} is a four-class (``\emph{World}'', ``\emph{Sports}'', ``\emph{Business}'', ``\emph{Sci/Tech}'') news topic classification dataset constructed by assembling titles and description fields of news articles.

\paragraph{Attack Methods}
To demonstrate the effectiveness of DPoE against various types of backdoor triggers, we choose three representative backdoor attack methods: word triggers, sentence triggers, and syntactic triggers.
(1) \textbf{BadNet} \cite{gu2017badnets} is originally proposed to attack image classification models.
We use the adapted version for text \cite{kurita-etal-2020-weight} that randomly inserts rare words as triggers.
(2) \textbf{InsertSent} \cite{dai2019backdoor} randomly inserts a fixed sentence as the backdoor trigger, for which we follow the default hyper-parameters in the original paper.
(3) \textbf{Syntactic} \cite{qi-etal-2021-hidden} is an invisible textual backdoor attack where syntactic structure is used as the trigger by
paraphrasing a victim sentence into the specified syntactic structure $S(SBAR)(,)(NP)(VP)(.)$.
Besides the attacks with a single type of triggers, we also propose a novel setting of (4) \textbf{Multi-Type} triggers where we mix all of the three types of triggers and insert one random type of trigger into each poisoned sample.

We use OpenBackdoor \cite{cui2022unified} for poisoned data generation. To be consistent with previous studies \cite{dai2019backdoor, qi-etal-2021-hidden, jin2022wedef}, we adopt a poison rate of $5\%$ for BadNet and InsertSent attack, and $20\%$ for syntactic and multi-type attack (mixing 10\%, 5\%, and 5\% of syntactic, BadNet, and InsertSent respectively).
We also show the defense results under different poison rates in \Cref{analysis:poison_rate}.


\paragraph{Baseline Methods}
\label{exp:baseline}
We compare our method DPoE with four representative defense methods.
(1) \textbf{ONION} \cite{qi-etal-2021-onion} detects and removes the suspicious words that are probably the backdoor triggers.
GPT-2 \cite{radford2019language} is used to evaluate the suspicion score of each word by the decrement of sentence perplexity after removing the word.
(2) \textbf{BKI} \cite{chen2021mitigating}, short for Backdoor Keyword Identification, detects trigger words and discards poisoned samples from the training data for purification.\footnote{Similarly, BKI leverages a scoring function to evaluate the importance of every single word to the model's prediction. A higher importance score indicates a higher probability for a word to be a trigger.}
(3) \textbf{STRIP} \cite{gao2021design} filters out poisoned samples by checking the inconsistency
of model's predictions when the input is perturbed several times.\footnote{The intuition is that it is difficult for any perturbation to the poisoned samples to influence the predicted class as long as the trigger exists.}
(4) \textbf{RAP} \cite{yang-etal-2021-rap} uses a fixed perturbation and a threshold of the output probability change of the protect label (decided by the defender) to detect poisoned samples in the inference stage.

\paragraph{Implementation and Evaluation Metrics}
\label{exp:imp}
To be consistent with previous study \cite{qi-etal-2021-onion, yang-etal-2021-rap, jin2022wedef}, we use BERT-base-uncased model \cite{devlin-etal-2019-bert} as the backbone of the DPoE framework.
We also report results on Llama-2-7B~\cite{touvron2023llama} to validate the effectiveness of the proposed algorithm on models of varying scales.
All experiments are conducted on a single NVIDIA RTX A5000 (for BERT-base-uncased) or RTX 8000 (for Llama-2-7B).
We train all models for $3$ epochs and pick the best hyper-parameter based on the pseudo development set strategy (\Cref{method:pseudo_dev}).
The defense methods are evaluated with the following two metrics.
(1) Clean accuracy (\textbf{Acc}) measures the performance of the defense model on the clean test data;
(2) Attack success rate (\textbf{ASR}) computes the percentage of trigger-embedded test samples that are classified as the target class by the defend model.
Following \citet{jin2022wedef}, we also demonstrate the results of \textbf{NoDefense} and \textbf{Benign} for a more comprehensive understanding on the performance of the defense mechanisms.
\textbf{NoDefense} is a vanilla BERT-base model fine-tuned on the poisoned data without any defense; \textbf{Benign} is a model trained on the clean data without poisoned samples.
These two baselines are either provided with full prior knowledge of the attack, or free of attack, representing ideal situations that are not accessible to a defense model.

\begin{table}[t]
\setlength\tabcolsep{1.5pt}
\renewcommand{\arraystretch}{1.3}
\centering
\footnotesize
\begin{tabular}{lcccccccc}
\hline \hline
\multicolumn{1}{l|}{\multirow{2}{*}{\textbf{Methods}}} & \multicolumn{2}{c}{\textbf{BadNet}} & \multicolumn{2}{c}{\textbf{InsertSent}} & \multicolumn{2}{c}{\textbf{Syntactic}} & \multicolumn{2}{c}{\textbf{Multi-Type}} \\ \cline{2-9} 
\multicolumn{1}{l|}{} & \textbf{ASR} & \textbf{Acc} & \textbf{ASR} & \textbf{Acc} & \textbf{ASR} & \textbf{Acc} & \multicolumn{1}{c}{\textbf{ASR}} & \multicolumn{1}{c}{\textbf{Acc}} \\ \hline \hline
\multicolumn{9}{c}{OffensEval} \\ \hline
\multicolumn{1}{l|}{NoDef.} & 96.77 & 84.05 & 99.84 & 83.93 & 98.71 & 83.93 & 93.62 & 82.40 \\
\multicolumn{1}{l|}{ONION} & 21.49 & 80.21 & 96.71 & 78.32 & 95.45 & 76.24 & 77.43 & 78.95 \\ \hline
\multicolumn{1}{l|}{DPoE} & \textbf{8.89} & \textbf{83.00} & \textbf{0.00} & \textbf{81.25} & \textbf{0.00} & \textbf{80.44} & \textbf{6.32} & \textbf{81.93} \\ \hline
\multicolumn{9}{c}{SST-2} \\ \hline
\multicolumn{1}{l|}{NoDef.} & 93.75 & 95.77 & 99.89 & 96.49 & 95.18 & 95.88 & 91.93 & 94.41 \\
\multicolumn{1}{l|}{ONION} & 26.98 & 89.31 & 98.90 & 87.84 & 94.82 & 83.26 & 87.57 & 84.29 \\ \hline
\multicolumn{1}{l|}{DPoE} & \textbf{7.32} & \textbf{94.63} & \textbf{15.26} & \textbf{94.89} & \textbf{19.54} & \textbf{93.67} & \textbf{16.33} & \textbf{93.85} \\ \hline \hline
\end{tabular}
\caption{Defense performance of DPoE with R-Drop on Llama 2. Best results are \textbf{boldfaced}. * NoDef. (NoDefense) is for reference and is not directly comparable with defense results.}
\label{tab:llama}
\end{table}

\subsection{Main Results}
\label{exp:main}
As shown in \Cref{tab:main}, our proposed DPoE method outperforms all of the four baselines and achieves the best defense performance on all of the three single-type trigger attacks as well as the muti-type trigger setting, especially the syntactic attack that most baseline methods fail to defend against.
Since ONION and BKI are detection-based defense methods based on the assumption that triggers are rare words, the syntactic attack which does not involve explicit trigger words is not within their scope of defense.
In contrast, DPoE leverages a shallow model to capture the backdoor shortcuts regardless of the type of triggers,\footnote{We demonstrate in \Cref{appendix:clean} that DPoE remains robust when there is no backdoor in the training data.} enabling the training of a backdoor-robust main model that does not learn the backdoor shortcut from triggers to the target label.
Furthermore, DPoE achieves an even lower ASR than the Benign baseline in some cases (highlighted in blue), indicating that DPoE not only effectively defends backdoor triggers, but is also robust to the semantic shortcuts introduced by the insertion of triggers.
More importantly, the clean Acc of DPoE can be higher than Benign model, enabling backdoor defense to no longer come at the expense of clean data accuracy.

\textbf{NoDefense} and \textbf{Benign} provide an understanding of the attack effectiveness and the defense performance.
The ASR of multi-type triggers exceeds $96\%$ on BERT without defense for all of three datasets, indicating the effectiveness of the mix trigger attack which can induce the victim model to predict the target label almost certainly.
Under the novel mix trigger attack, DPoE also manages to defend effectively with the ASR being close to or even lower than that of Benign.
Besides, DPoE still maintains a competitive clean Acc compared with Benign and NoDefense, which demonstrates the effectiveness of the denoising technique that helps the model to be more resistant to noisy labels even under a high noise rate (approximately $20\%$ for syntactic and multi-type attacks).

Compared with applying PoE alone, the clean Acc is significantly improved due to the denoising module of DPoE. On OffensEval, Acc under BadNet attack is only $81.72\%$ by PoE defense, while exceeding $84\%$ after applying the denoising technique, which performs even better than the Benign model.
This is because the benign training data might already contain noisy labels that hinder the utility of the model, which is alleviated by the denoising module in DPoE.
Similar conclusions can be made on all of the three datasets under four attacks.
Overall, the incorporated denoising part further boosts defense performance, while no single denoising technique consistently outperforms the rest.
The most effective denoising scheme for backdoor defense is left for future work.

To validate its effectiveness on models of larger scales, we also apply DPoE with R-Drop to Llama-2-7B. As illustrated in \Cref{tab:llama}, DPoE outperforms the ONION baseline and achieves
highly competitive defense performance under all of the four different types of attacks on both SST-2 and OffensEval datasets. For instance, DPoE maintains an ASR of below 10\% on the OffensEval dataset under all of the four attack settings, while ONION fails to take effect under sentence- and syntactic-level attacks with ASR exceeding 95\%. Similar observations can be found on the SST-2 dataset. As a result, DPoE remains effective along with the scale-up of the backbone language model, indicating its robustness in handling larger and more complex backbone architectures without significant loss in performance or efficiency.



\begin{figure}[t]
	\centering
	\subfigure[No Defense]{
		\centering
		\includegraphics[width=7.5cm]{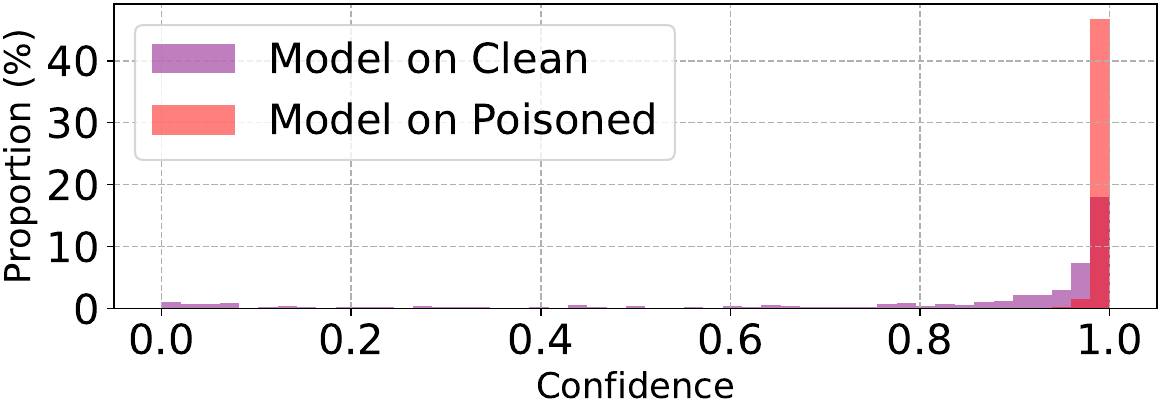}
        \label{fig:logits_no_defense}
	}
	\subfigure[DPoE w/ R-Drop]{
		\centering
		\includegraphics[width=7.5cm]{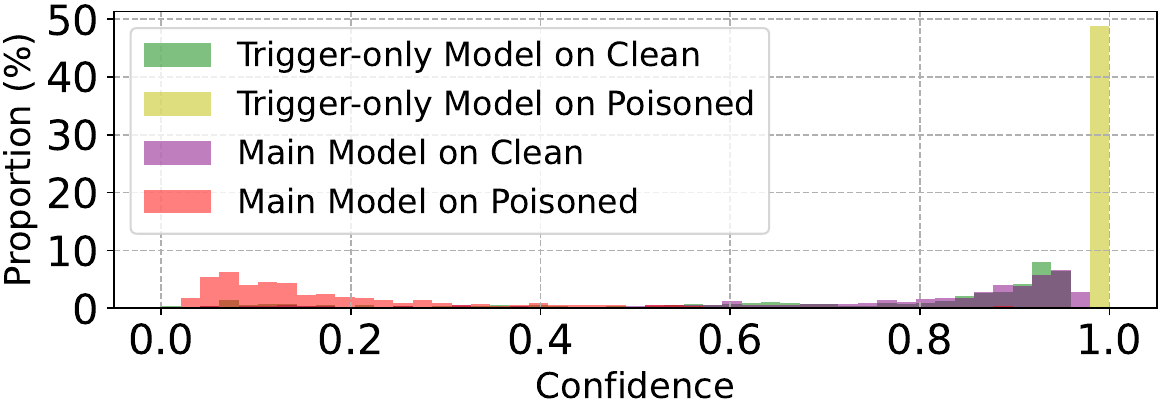}
        \label{fig:logits_rdrop}
	}
 \vspace{-0.5em}
	\caption{Prediction confidence distribution with (bottom) and without (top) DPoE defense. DPoE results in high confidence of trigger-only model on poisoned samples, enabling a backdoor-resistant robust main model.}
	\label{fig:logits_distribution}
 \vspace{-1em}
\end{figure}

\subsection{Analysis}
\label{exp:analysis}

\stitle{Effect of DPoE for Defense}
\label{analysis:logits_distribution}
To understand the influence of DPoE on the trigger-only model and the main model, we examine the confidence distribution of BERT without defense (\Cref{fig:logits_no_defense}) and with DPoE (\Cref{fig:logits_rdrop}) when trained on OffensEval poisoned by syntactic trigger at $20\%$ poison rate.
BERT without defense undoubtedly learns the backdoor shortcut and predicts most of the poisoned samples as the target label with almost $100\%$ confidence.
In contrast, the trigger-only model captures the backdoor shortcut and also predicts poisoned samples with high confidence, leaving the main model with trigger-free residual so that the main model learns to assign rather low confidence on poisoned data.
This change in confidence distribution reveals the inner influence of DPoE for effectively preventing main model from learning the shortcut from backdoor triggers to target label.


\begin{figure}[t]
	\centering
	\subfigure[ASR]{
		\centering
		\includegraphics[width=3.5cm]{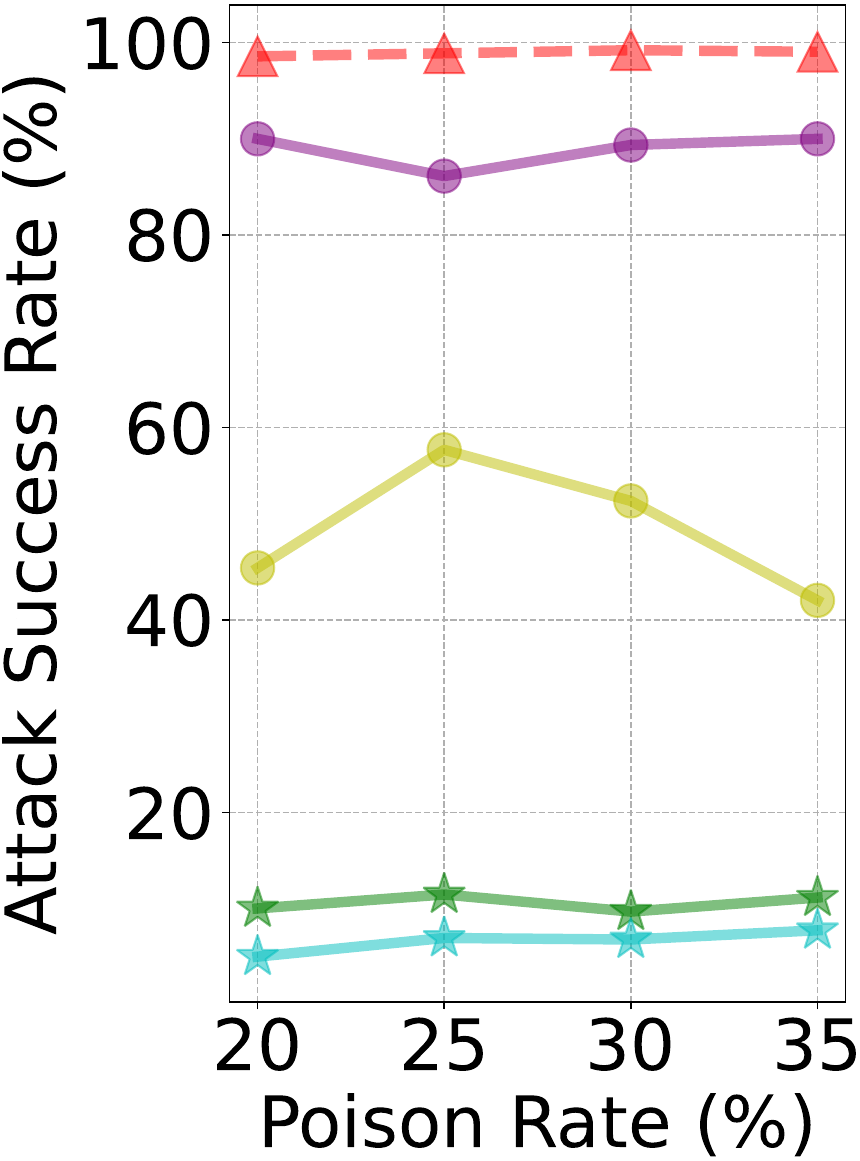}
		\label{fig:poison_rate_asr}
	}
	\subfigure[Acc]{
		\centering
		\includegraphics[width=3.5cm]{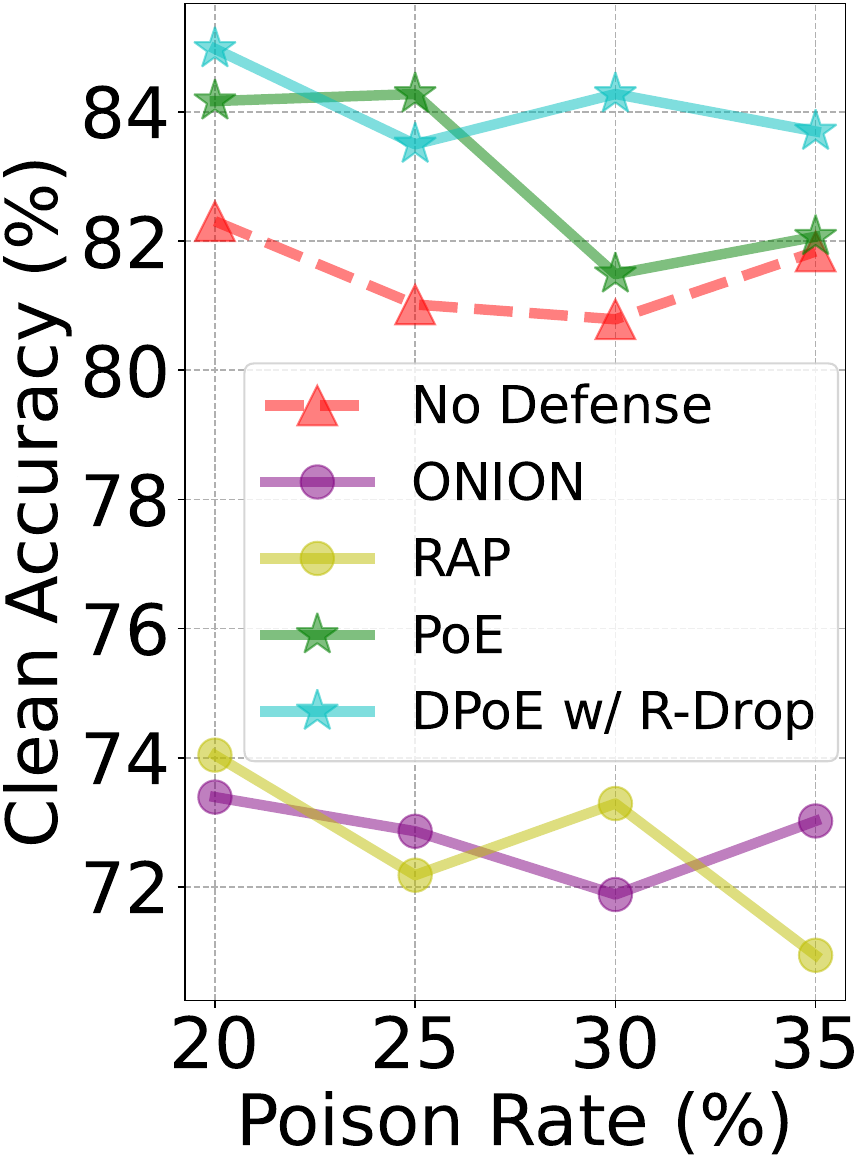}
		\label{fig:poison_rate_acc}
	}
 \vspace{-0.5em}
	\caption{Attack success rate (left) and clean accuracy (right) of defense methods by different poison rates on OffensEval task under syntactic attack. DPoE maintains competitive defense performance as poison rate rises.}
	\label{fig:ablation_poison_rate}
 \vspace{-1em}
\end{figure}



\stitle{Higher Poison Rate}
\label{analysis:poison_rate}
To examine the resistance of DPoE against more devastating attacks, we challenge it with higher poison rate on the OffensEval task under syntactic attack.
\Cref{fig:poison_rate_asr} shows that, for all the listed defense methods, there is not much increase in the ASR with the rise of poison rate, indicating that poison rate of $20\%$ is enough for the victim model to be poisoned and sufficiently learn the backdoor shortcut.
This phenomenon is consistent with previous study \cite{qi-etal-2021-hidden}.
Though the ASR is not much affected by higher poison rate, there is an obvious decrease on the clean accuracy of baseline methods (ONION and RAP, \Cref{fig:poison_rate_acc}).
The decrease in model utility is due to the fact that higher poison rate brings about more noisy labels, hindering the model from learning task-relevant features.
In contrast, DPoE maintains stable performance of clean accuracy due to the denoising mechanism, indicating that DPoE remains competitive against more challenging attacks.

\section{Conclusion}
In this paper, we propose DPoE, an end-to-end ensemble-based backdoor defense method that mitigates backdoor triggers by learning the backdoor-free residual of a shallow model that captures the backdoor shortcuts.
In addition to debiasing-based trigger mitigation and denoising techniques,
a pseudo development set construction strategy is also proposed for hyper-parameter tuning since a clean dev set is absent in real-world scenarios.
Experiments on three NLP tasks demonstrate its effectiveness in defending against various backdoor triggers as well as mix types of triggers.

\section*{Limitations}

The current investigation of DPoE has the following limitations.
First, although our experiments follow the settings of previous works for a fair comparison, the experimented tasks, types of triggers, languages, and backbone models can be further increased.
Since our framework is model-agnostic, 
experimentation with more backbone language models can be conducted, which we leave as future work to due to limited bandwidth.
Second, while we evaluate our method on 
discriminative NLU tasks
to align with previous studies, 
the proposed method has the potential to be extended for generative tasks similarly as the contrastive decoding method \cite{li2022contrastive}.
However, non-trivial adaption and systematic study will be needed to achieve this goal.
Third, DPoE applies only to training time defense which assumes that the defender has access to the training phase of a model.
We leave inference time defense for black-box models to future work.




\section*{Ethics Statement}



In this paper, we propose a defense method against backdoor attacks with different types of triggers.
Experimenting on three datasets that are publicly available, we show that our defense method effectively alleviates backdoor attacks without any prior knowledge about the backdoor triggers.
Therefore, our framework provides an efficient solution to potential misuse of language models and protects models from malicious attacks.
Besides, we also reveal one more adverse scenario of backdoor attack where various types of triggers are mixed together, disabling previous trigger-detection-based defense methods that assume the triggers to be rare words only.
We would like to raise researchers' attention towards this potential risk and call for defense methods that can be universally adapted against various trigger types.
Overall, the energy we consume for running the experiments is limited. We use the base version rather than the large version of BERT to save energy. No demographic or identity characteristics are used in this paper.

\section*{Acknowledgement}

Qin Liu and Muhao Chen were supported by the NSF Grant IIS 2105329, the NSF Grant ITE 2333736, the DARPA AIE Grant HR0011-24-9-0370, the Faculty Startup Fund of UC Davis and an Amazon Research Award.
Fei Wang was supported by the Amazon ML Fellowship.
Chaowei Xiao was supported by the U.S. Department of Homeland Security under Grant 17STQAC00001-06-00.


\bibliography{anthology,custom}
\bibliographystyle{acl_natbib}

\appendix


\begin{center}
    {\Large\textbf{Appendices}}
\end{center}

\section{Justification of PoE for Defense}
\label{append:justification}
Probability of label $y_i$ for example $x_i$ in the PoE ensemble is computed as
\begin{equation*}
    \hat{p}_{iy_i} = \sigma(\log(p_{iy_i} \cdot b_{iy_i})) = \frac{p_{iy_i} \cdot b_{iy_i}}{\sum_{k=1}^{|\mathcal{Y}|}p_{ik} \cdot b_{ik}},
\end{equation*}
where $\sigma$ denotes the softmax function.
Then the gradient of the CE loss $\mathcal{L}(\theta_h;\theta_g)$ w.r.t. $\theta_g$ is \cite{karimi-mahabadi-etal-2020-end}:
\begin{equation*}
\begin{aligned}
    \nabla_{\theta_g} \mathcal{L}(\theta_h;\theta_g) &= \\
    - \frac{1}{N} \sum_{i=1}^N &\sum_{k=1}^{|\mathcal{Y}|} \left[ \left(\delta_{y_ik} - \hat{p}_{ik}\right) \nabla_{\theta_g} \log(p_{ik}) \right],
\end{aligned}
\end{equation*}
where $\delta_{y_ik}$ equals $1$ when $k = y_i$ otherwise $0$.
Generally speaking, when both the trigger-only model and the main model have captured the backdoor associations, $\hat{p}_{ik}$ would be close to $1$ so that $(\delta_{y_ik}-\hat{p}_{ik})$ is close to $0$, decreasing the gradient of sample $i$.
On the contrary, when the sample is trigger-free, the trigger-only model predicts the uniform distribution over all classes $b_{ik} \approx \frac{1}{|\mathcal{Y}|}$ for $k \in \mathcal{Y}$. Therefore, $\hat{p}_{iy_i} = p_{iy_i}$ and the gradient of PoE classifier remains the same as CE loss.

\begin{figure}[t]
	\centering
	\subfigure[Clean Accuracy]{
		\centering
		\includegraphics[width=7.5cm]{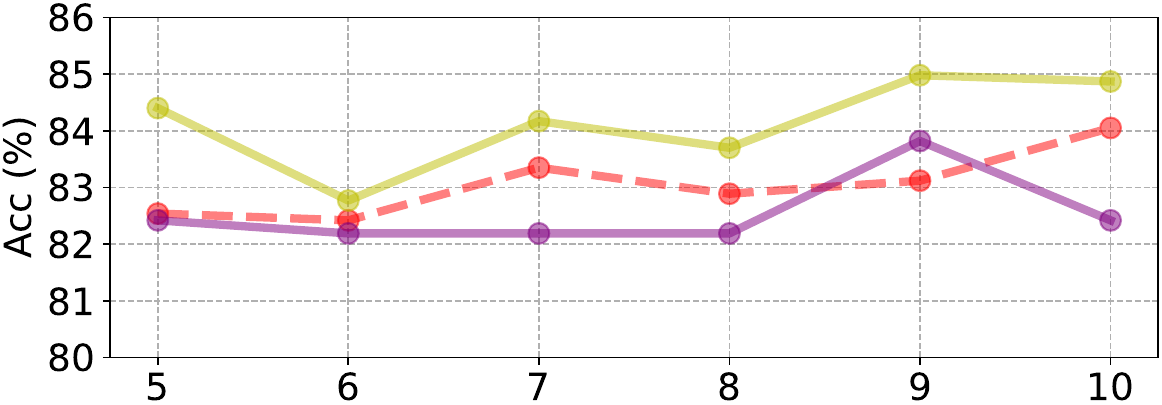}
        \label{fig:alpha_acc}
	}
	\subfigure[Attack Success Rate]{
		\centering
		\includegraphics[width=7.5cm]{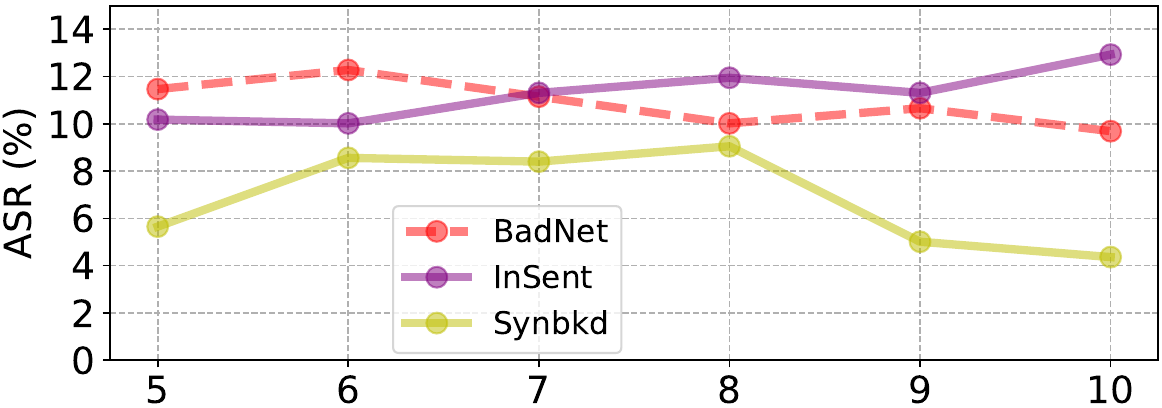}
        \label{fig:alpha_asr}
	}
 \vspace{-0.5em}
	\caption{Performance of DPoE by PoE coefficient $\beta$ on OffensEval task against three backdoor attacks. DPoE is steadily effective within a reasonable range of hyper-parameter values.}
	\label{fig:alpha}
 \vspace{-1em}
\end{figure}

\section{Validity of Pseudo Development Set}
\label{sec:append_pseudo_valid}

We denote the poison rate, true clean accuracy of the main model on clean test data, and accuracy on pseudo clean dev set as $\alpha = |\mathcal{D}^*| / |\mathcal{D}|$, $acc$, and $acc^*$ respectively.
Assume the poison rate of the selected pseudo poisoned dev set $\mathcal{D}_P$ and pseudo clean dev is $\alpha_p$ and $\alpha_c$ respectively, the real attack success rate for the main model is $asr$ (refer to the definition of metrics in \Cref{exp:imp}), and the accuracy on $\mathcal{D}_P$ is $asr_p$.

Firstly, suppose the main model performs well on both the pseudo poisoned dev set and the poisoned training set, which is equivalent to low $asr_p$ and high $acc^*$, it indicates low $asr$ and high $acc$:
\begin{equation*}
\begin{aligned}
    acc^* &= (1-\alpha_c) * acc + \alpha_c * asr, \\
    asr_p &= (1-\alpha_p) * acc + \alpha_p * asr.
\end{aligned}
\end{equation*}
Due to the high-accuracy low-recall strategy, $\alpha_c * asr$ can be ignored since $\alpha_c$ is close to zero and we have $acc^* \propto acc$.
On the other hand, a well-trained trigger-only model results in high $\alpha_p$ so that $asr_p \propto asr$.
So we have demonstrated that we can infer from the low $asr_p$ and high $acc^*$ that the main model is effective for defense.

Secondly, suppose there exists a main model with high $acc$ and low $asr$, which means it is an effective defense model indeed. Similarly, we have:
\begin{equation*}
\begin{aligned}
    acc &= \frac{acc^* - \alpha_c * asr}{1-\alpha_c}, \\
    asr &= \frac{asr_p - (1-\alpha_p) * acc}{\alpha_p}.
\end{aligned}
\end{equation*}
When the poison rate $\alpha_c$ is low, $(1-\alpha_c) \approx 1$ so that $acc \propto acc^*$.
Further, an ideal main model indicates an effective trigger-only model that performs high $\alpha_p$, which means $asr \propto asr_p$.
So we have illustrated that a promising main model will be detected and selected by the pseudo dev set.
As a result, our construction of pseudo dev set is valid since $asr_p$ and $acc^*$ on the pseudo dev set are effective approximations of $asr$ and $acc$.

\section{Effect of PoE Coefficient}
\label{appendix:beta}
The coefficient $\beta$ in \Cref{eq:poe} denotes the weight we assign to the predicted probability distribution of the trigger-only model in the framework of PoE.
To examine whether our defense strategy is sensitive to this hyper-parameter, we evaluate DPoE with different $\beta$ coefficient on the OffensEval task under three types of backdoor attacks.
As shown in \Cref{fig:alpha}, the overall performance of DPoE slightly fluctuates with different coefficients, while indicating that DPoE remains effective within a reasonable range of hyper-parameter values.

\section{DPoE on Clean Dataset}
\label{appendix:clean}
To further examine the validity of DPoE, we train the model on the clean datasets of three tasks.
Clean accuracy shown in \Cref{tab:clean} proves that DPoE does not hurt normal performance when being trained without triggers for the shallow model to capture. In this case, the trigger-only model learns superficial features (spurious correlations) that are no more desired than the trigger-related feature(s) and a robust main model should not make predictions based on these shallow features \cite{gardner-etal-2021-competency}. Thus, the trigger-only model’s learning of shallow features would help the main model mitigate these shallow-feature-related spurious correlations and further boost its performance and robustness with the help of PoE.

\begin{table}[t]
\setlength\tabcolsep{5pt}
\centering
\small
\begin{tabular}{l|ccc}
\hline \hline
\textbf{Methods} & \textbf{SST-2} & \textbf{OffensEval} & \textbf{AG News} \\ \hline
\rowcolor{gray!20} 
Finetune & 91.16 & 83.47 & \underline{94.49} \\ \hline
PoE & 91.27 & 83.29 & 94.32 \\ \hline
DPoE w/ R-Drop & \textbf{91.76} & \textbf{85.10} & 94.37 \\
DPoE w/ LS & 91.49 & 83.70 & 94.41 \\
DPoE w/ Re-weight & \underline{91.54} & 83.93 & 94.21 \\
DPoE w/ SL & 91.43 & \underline{84.98} & \textbf{94.89} \\ \hline \hline
\end{tabular}
\caption{Clean accuracy of DPoE trained on clean datasets. The best results are \textbf{boldfaced} and the second best are \underline{underlined}.}
\label{tab:clean}
\end{table}

\end{document}